\documentclass[conference]{IEEEtran}
\usepackage{times}
\usepackage[numbers]{natbib}
\usepackage{multicol}
\usepackage{booktabs}
\usepackage{multirow}
\usepackage{xcolor}

\newcommand{\myparagraph}[1]{\vspace{0.05in}\noindent\textbf{#1}}
\definecolor{MyDarkBlue}{rgb}{0,0.08,1}
\definecolor{MyDarkGreen}{rgb}{0.02,0.6,0.02}
\definecolor{MyDarkRed}{rgb}{0.8,0.02,0.02}
\definecolor{MyDarkOrange}{rgb}{0.40,0.2,0.02}
\definecolor{MyPurple}{rgb}{111,0,255}
\definecolor{MyRed}{rgb}{1.0,0.0,0.0}
\definecolor{MyGold}{rgb}{0.75,0.6,0.12}
\definecolor{MyDarkgray}{rgb}{0.66, 0.66, 0.66}

\newcommand{\fig}[1]{Fig.~\ref{#1}}
\newcommand{\eq}[1]{Equation~\eqref{#1}}

\newcommand{\tab}[1]{Table~\ref{#1}}

\usepackage{graphics}
\usepackage{epsfig}
\usepackage{mathptmx}
\usepackage{times}
\usepackage{amsmath}
\usepackage{amssymb}
\usepackage{color}
\usepackage{textcomp}
\usepackage{gensymb}
\usepackage{float}
\usepackage{amsmath}
\usepackage[pagebackref=true,breaklinks=true,colorlinks,bookmarks=false]{hyperref}

\newcommand{\sensor}[0]{DTact}

\begin{document}

\title{\LARGE \bf
\sensor{}: A Vision-Based Tactile Sensor that Measures High-Resolution 3D Geometry Directly from Darkness
}

\author{
    \authorblockN{Changyi Lin$^1$, Ziqi Lin$^2$, Shaoxiong Wang$^3$, Huazhe Xu$^{1, 2}$}
    \authorblockA{
     $^1$Shanghai Qi Zhi Institute, $^2$Tsinghua University, $^3$Massachusetts Institute of Technology\\
    {\tt\small linchangyi1@foxmail.com, huazhe\_xu@mail.tsinghua.edu.cn}
}
    \href{https://sites.google.com/view/dtact-sensor}{https://sites.google.com/view/dtact-sensor}
}

\maketitle


\begin{abstract}
Vision-based tactile sensors that can measure 3D geometry of the contacting objects are crucial for robots to perform dexterous manipulation tasks. However, the existing sensors are usually complicated to fabricate and delicate to extend. In this work, we novelly take advantage of the reflection property of semitransparent elastomer to design a robust, low-cost, and easy-to-fabricate tactile sensor named \sensor{}. \sensor{}  measures high-resolution 3D geometry accurately from the darkness shown in the captured tactile images with only a single image for calibration. In contrast to previous sensors, \sensor{} is robust under various illumination conditions. Then, we build prototypes of \sensor{} that have non-planar contact surfaces with minimal extra efforts and costs. Finally, we perform two intelligent robotic tasks including pose estimation and object recognition using \sensor{}, in which \sensor{}  shows large potential in applications.
\end{abstract}


\section{Introduction}
\label{sec:intro}

Tactile perception is essential for robots to sense and interact with the physical world. It enables robots to perceive rich physical information of contact objects~\cite{abad2020visuotactile} such as 3D geometry, force, slip, hardness, and temperature. Among the tactile information, 3D geometry stands out for its ability to infer shapes, poses, textures, and normal forces. Therefore, a tactile sensor that measures 3D geometry accurately and robustly is desired for downstream tasks such as robotic manipulation.

To satisfy the requirement, vision-based tactile sensors such as GelSight are developed to measure high-resolution 3D geometry of the contact objects~\cite{yuan2017gelsight}. 
Despite the success of these sensors, the core photometric stereo technique~\cite{johnson2009retrographic} has strict requirements for the directions and uniformity of internal illumination. The delicacy significantly increases the complexity of mechanical design and the difficulty of assembly to construct a functional sensor. Besides, such high dependency on illumination strongly hinders extension of GelSight-style sensors into non-planar shapes, which results in many existing sensors~\cite{dong2017improved, donlon2018gelslim, ma2019dense, lambeta2020digit, wang2021gelsight, taylor2022gelslim} having flat surfaces.

\begin{figure}[t]
	\centering
	\includegraphics[width= \linewidth]{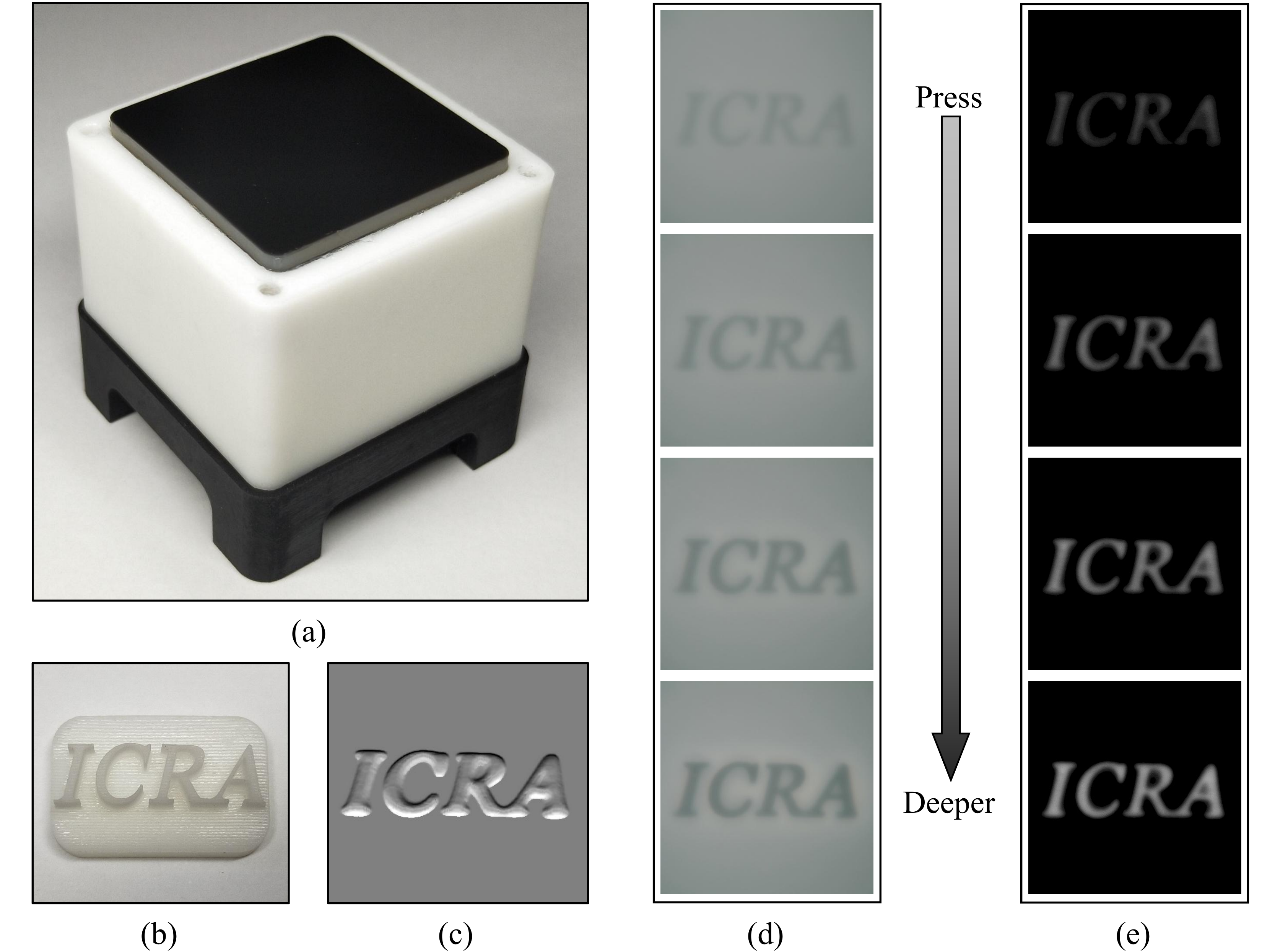}
	\caption{(a) The \sensor{} sensor. (b) The 3d-printed badge used to press. (c) Reconstruction result of the badge with \sensor{}. (d) The contact areas show darker when they are pressed deeper. (e) Depth maps corresponding to the captured images in (d). }
	\label{fig:abstract}
\end{figure}

In this work, we creatively leverage the reflection property of the semitransparent elastomer to form a new combination of elastomer layers. With external illumination only, the region of these layers that is pressed deeper would result in a darker color when captured by a camera, as shown in~\fig{fig:abstract} (d). Employing these layers as the sensing surface, we propose \sensor{}, a robust, low-cost, and easy-to-fabricate tactile sensor that measures high-resolution 3D geometry accurately with a monocular camera. Thanks to the simple yet effective design, \sensor{} only needs a single image for intensity-to-depth calibration; the depth map is directly mapped from the grayscale value. Furthermore, \sensor{} can be easily extended to non-planar sensors because of its low dependency on illumination.

We evaluate the performance of \sensor{} in shape reconstruction from both quantitative and qualitative perspectives. Then we test its robustness under various illumination conditions and explore the effect of elastomer thickness in measuring fine geometry. We also demonstrate its large potential to be extended to  sensors with non-planar surfaces. Lastly, we use \sensor{} for two downstream applications: object pose estimation and object recognition.

The rest of this paper is organized as follows. We introduce related work on vision-based tactile sensors for measuring 3D geometry in Section~\ref{sec:related}. We describe the design and fabrication of \sensor{} in Section~\ref{sec:sensor}. We then present the methods, implementation, and results of shape reconstruction in Section~\ref{sec:reconstruction}. We evaluate \sensor{}'s robustness and potential from different aspects in Section~\ref{sec:experiments}. Next, we apply \sensor{} to perform two downstream tasks including pose estimation and object recognition in Section~\ref{sec:application}. Finally, we give the conclusion in Section~\ref{sec:conclusion}.

\section{Related Work}
\label{sec:related}

Vision-based tactile sensors~\cite{yuan2017gelsight, ward2018tactip} have become increasingly popular in the robotics community in recent years. GelSight sensor~\cite{yuan2017gelsight} and various revised ones~\cite{dong2017improved, wang2021gelsight, taylor2022gelslim, lepora2022digitac} use the photometric stereo technique~\cite{johnson2009retrographic} to measure high-resolution 3D geometry of contact objects with a monocular camera. With a GelSight sensor mounted on the gripper, robots can accomplish challenging tasks such as texture recognition~\cite{li2013sensing, luo2018vitac, yuan2018active}, dexterous manipulation~\cite{tian2019manipulation,wang2020swingbot, she2021cable, dong2021tactile}, shape mapping~\cite{bauza2019tactile,suresh2022shapemap}, as well as liquid property estimation~\cite{huang2022understanding}.

However, photometric stereo technique has strict requirements for the light propagation paths. Specifically, GelSight sensor has to arrange three colors of light~(RGB) in different directions to provide internal illumination for a transparent elastomer layer, so that it can use a look-up table to map the changing RGB values to gradients for every pixel. The aforementioned technique  brings the sensors high dependency on internal illumination, which further complicates the fabrication process including precise arrangements for LEDs and fine finishing of the silver pigment layer~\cite{yuan2017gelsight}. These problems invoke considerable hurdles to construct a GelSight-like sensor without sufficient expertise, hindering the reproduction of these sensors in the robotics community~\cite{du2021high}. Besides, it is hard for these sensors to work under large deformation or measure a slab with zero gradient everywhere.

To resolve these issues, researchers explore the possibilities to build tactile sensors for measuring 3D geometry without using photometric stereo technique. DelTact~\cite{zhang2022deltact} uses optical flow to estimate the deformation of a dense color pattern, which reduces the dependency on illumination. However, the reconstruction results of DelTact often suffer from the artifacts caused by the estimation algorithms. Moreover, producing the required color patterns is usually time-consuming. Contrastingly, other works directly replace the monocular camera by specialized hardware. Soft-bubble~\cite{alspach2019soft} uses a depth sensor to sense the shape of an air-filled membrane, which increases the size and cost of the sensor. GelStereo~\cite{cui2021hand} applies a binocular camera to capture image pairs for sparse surface reconstruction. Improved with self-supervised neural networks~\cite{cui2021self}, GelStereo is able to measure dense 3D geometry at the cost of training neural networks and occasional inaccurate measures. In general, these sensors compromise on certain aspects such as the accuracy, the size, the cost, or the algorithmic complexity to reduce the dependency on illumination. By contrast,  \sensor{} measures high-resolution 3D geometry accurately with a monocular camera under simple external illumination. The fabrication, calibration, and reconstruction processes are also improved and simplified in comparison to other sensors.

In real-world robotic tasks, the robots usually require different non-planar shapes of tactile sensor surfaces, to fit either a specialized task, or the mechanical design of their end effector.
While some non-planar sensors exist, they may require high fabrication cost~\cite{romero2020soft, padmanabha2020omnitact} or large amounts of data for training the reconstruction models~\cite{do2022densetact, sun2022soft}. Unlike previous works, \sensor{} can be extended with a non-planar surface while keeping the illumination condition and reconstruction algorithm the same as planar ones. We attribute these advantages to its robustness against the illumination.


\section{Design and Fabrication}
\label{sec:sensor}

\begin{figure}[t]
	\centering
	\includegraphics[width= \linewidth]{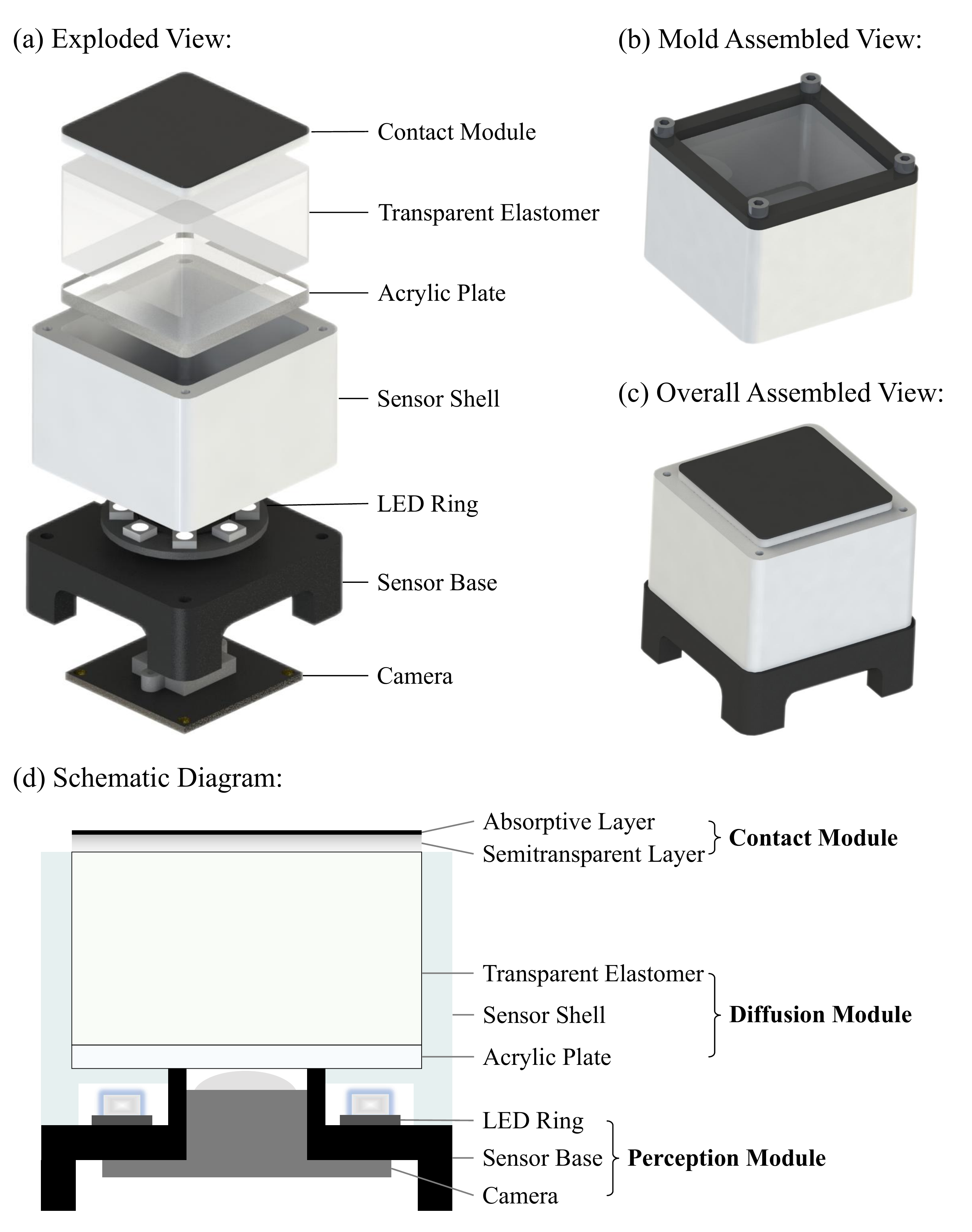}
	\caption{Design of \sensor{}. (a) Components of \sensor{} are shown in exploded view. (b) The open mold in black is locked on the sensor shell. (c) The assembled view of \sensor{}. (d) The schematic diagram of \sensor{}.}
	\label{fig:design}
\end{figure}

\begin{figure}[t]
	\centering
	\includegraphics[width= \linewidth]{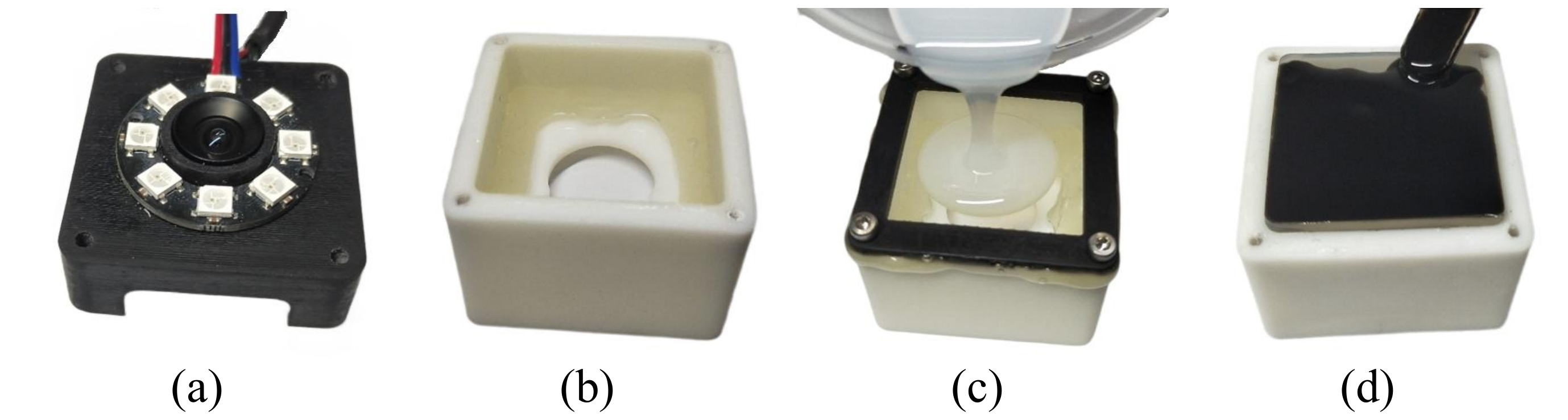}
	\caption{Four main fabrication steps of \sensor{}. (a) Fabrication of the perception module. (b) Fabrication of the diffusion module. (c) The resinic mold is fastened to the diffusion module with M2 screws and the semetransparent silicone is poured into it. (d) The black silicone is painted on the semitransparent layer.}
	\label{fig:fabrication}
\end{figure}

In this section, we aim to elucidate the design and fabrication process of the \sensor{} tactile sensor. We propose a new combination of elastomer layers which directly maps the depth of the elastomer to the intensity of an image. The main components as well as the mold are existing commodities or manufactured by 3D printing as shown in~\fig{fig:design} (a). In contrast to previous holistic sensors such as~\cite{sferrazza2019design}, \sensor{} is modularized based on their structures and functions~(as shown in~\fig{fig:design} (d)): the perception module, the diffusion module, and the contact module. As for fabrication, it takes four steps to build the sensor~(as shown in~\fig{fig:fabrication}). Detailed descriptions about the modules and fabrication are presented as follows.

\myparagraph{The perception module.}
The perception module consists of a sensor base, a camera, and a LED ring.
The 3D-printed sensor base supports the whole sensor. We choose a USB camera with a high FOV of $120$ degrees. The camera captures image frames with a resolution of $800\times600$ at $60$ FPS. \sensor{} advocates the USB-based camera port to support multiple platforms, rather than using a Raspberry Pi camera as in previous works~\cite{donlon2018gelslim, wang2021gelsight, taylor2022gelslim, romero2020soft}. 
As for illumination, we use an off-the-shelf commercial LED ring~(WS2812B-8) to provide stable white light.

The camera is fastened to the sensor base with M2 screws, while the LED ring is nested in the center of the sensor base, as shown in~\fig{fig:fabrication} (a).

\myparagraph{The diffusion module.}
The diffusion module includes a sensor shell, an acrylic plate, and a block of transparent elastomer.
The sensor shell is 3D-printed with white resin and serves as the container for the transparent elastomer. Cut by laser, the 3-mm thick acrylic plate provides a clear window for the camera and supports the transparent elastomer. In order to avoid over-reflection from the LEDs, the transparent elastomer diffuses the light. Specifically, we use stiff transparent silicone~(ELASTOSIL\textsuperscript{\textregistered} RT 601, 9:1 ratio, shore hardness 45A) as the material of the transparent elastomer.

We first fix the acrylic plate to the sensor shell with hot-melt adhesive, as shown in~\fig{fig:fabrication} (b). Next, we remove the air bubbles in the mixed  transparent silicone with a vacuum pump. Finally, we pour it directly into the sensor shell. It takes about 24 hours for the silicone to cure at room temperature.

\myparagraph{The contact module.}
The contact module is made up of a semitransparent layer and an absorptive layer.
The soft semitransparent silicone~(POSILICONE\textsuperscript{\textregistered} , 1:1 ratio, shore hardness 5A which resembles that of human skin) is used to make the semitransparent layer whose brightness changes when deformation occurs. The thickness of the semitransparent layer is $2\texttt{mm}$. We then make an absorptive layer by mixing the same silicone with black silicone pigment. This layer absorbs the LED light going through the semitransparent layer, so that the tactile images only depends on the light reflected by the semitransparent layer. In addition, the absorptive layer also blocks light from outside environment. The areas that are pressed would become thinner than before. With the combination of the semitransparent layer and the absorptive layer, less light would be reflected by the semitransparent layer; the light going through it would be absorbed by the absorptive layer. Consequently, the contact areas become darker than before. This forms the basic principle of \sensor{}: the deeper an area is pressed, the darker it results in the output.

As shown in~\fig{fig:design} (b), we improve the production process of the contact module~\cite{sferrazza2019design} by locking an 3D-printed open mold on the sensor shell, into which the defoamed semitransparent silicone is poured directly~(\fig{fig:fabrication} (c)). This improvement can avoid dust and air bubbles between the semitransparent layer and the diffusion elastomer because the semitransparent layer never moves. After the semitransparent silicone cures, we use a stick to paint the defoamed black silicone on the surface of the semitransparent layer, as shown in~\fig{fig:fabrication} (d).

\fig{fig:design} (c) shows the result of assembling three modules together. The size of the sensor is $45\texttt{mm}\times45\texttt{mm}\times47\texttt{mm}$, which can be easily reduced by using smaller cameras and LED rings if downstream applications~(e.g., robotic manipulation) demand. The components of the sensor and the mold cost less than \$7 in total without the camera. The fabrication and assembly steps can be reproduced following the illustrated steps.


\section{3D Shape Reconstruction}
\label{sec:reconstruction}

\newcommand{\PixelPosition}{(u, v)}
\newcommand{\SurfacePosition}{(x, y)}
\newcommand{\InitialValue}{I_0(u, v)}
\newcommand{\CurrentValue}{I(u, v)}
\newcommand{\ChangedValue}{I_\Delta(u, v)}
\newcommand{\PressedDepth}{D(x, y)}
\newcommand{\MappingList}{M}
\newcommand{\LookupTable}{R}
\newcommand{\slops}{k(u, v)}

In this section, we introduce the methods and implementation details of the core function of \sensor: 3D shape reconstruction. We propose two possible methods for shape reconstruction in Section~\ref{sub:methods} and show the corresponding calibration results in Section~\ref{sub:calibration}. Then we describe the implementation details of reconstruction in Section~\ref{sub:reconstruction_details}. Finally, we evaluate these methods and examine the visual reconstruction results of representative objects in Section~\ref{sub:reconstruction_results}.

\subsection{Reconstruction Methods}
\label{sub:methods}
We aim to compute the 3D geometry from the pixel intensity of the captured images. Specifically, we need to calculate the pressed depth $\PressedDepth$ from the intensity variation value $\ChangedValue$ for every pixel, where $\PixelPosition$ is the location of a pixel in the image and $\SurfacePosition$ is the corresponding location on the sensing surface.

The ideal method to reconstruct the pressed depth is to build a look-up table $\LookupTable$ that takes as inputs the intensity variation value $\ChangedValue$ and the pixel's position $\PixelPosition$:
\begin{equation}
\PressedDepth = \LookupTable(\ChangedValue, u, v)\label{idea}
\end{equation}
However, it requires a high-precision computer numerical control~(CNC) machine to build the high dimensional look-up table, because precise continuous depth for every pixel is needed. This calibration procedure is usually time-consuming and tool-dependent. To simplify the calibration procedure, we propose the following two reconstruction methods in which parameters can be obtained without using a CNC machine.

\myparagraph{The single image method.} We simplify the look-up table $\LookupTable$ in~\eq{idea} into a mapping list $\MappingList$, which only takes the intensity variation value $\ChangedValue$ as input. For all pixels in the image, the pressed depth can be sought from the mapping list:
\begin{equation}
\PressedDepth = \MappingList(\ChangedValue)\label{single}
\end{equation}

\myparagraph{The linear regression method.} In order to take the illumination uniformity into account, we fit a linear mapping function for each pixel. Hence, each pixel would have a specific slope $\slops$:
\begin{equation}
\PressedDepth = \slops \cdot \ChangedValue \label{linear}
\end{equation}
Besides, the reference images captured by \sensor{} are slightly brighter in center owing to the circular distribution of the eight LEDs. Therefore, we fit another linear mapping function between the distance to center and the slope $\slops$:
\begin{equation}
\slops = k_c \times \sqrt{(u-u_c)^2+(v-v_c)^2} + b_c \label{center_fit}
\end{equation}
where $(u_c, v_c)$ is the center of the image.

Based on~\eq{linear} and~\eq{center_fit}, the pressed depth \PressedDepth  is finally modeled as:
\begin{equation}
\PressedDepth = (k_c \times \sqrt{(u-u_c)^2+(v-v_c)^2} + b_c) \times \ChangedValue \label{regression}
\end{equation}

\subsection{Calibration}
\label{sub:calibration}

In this section, we aim to calibrate \sensor{} to solve the parameters used for reconstruction, including $\MappingList$ in the single image method, $k_c$ and $b_c$ in the linear regression method. We press a metal ball on the \sensor{} and record the images. Knowing the radius of the pressed ball and the corresponding surface size of a pixel, we compute the actual depth maps for the collected images with circle detection algorithms in OpenCV~\cite{bradski2000opencv}. Therefore, we can infer the parameters from the pressed depth $\PressedDepth$ in the actual depth maps and their corresponding pixel intensity variation value $\ChangedValue$.

There are four channels of the pixel intensities can be used to represent $\ChangedValue$, including the RGB and grayscale channels. According to the empirical results of the noise levels of the four channels in the reference images, the grayscale channel has the most stable pixel intensity. Therefore, we choose grayscale channel for calibration and future downstream tasks.

\begin{figure}[t]
	\centering
	\includegraphics[width= \linewidth]{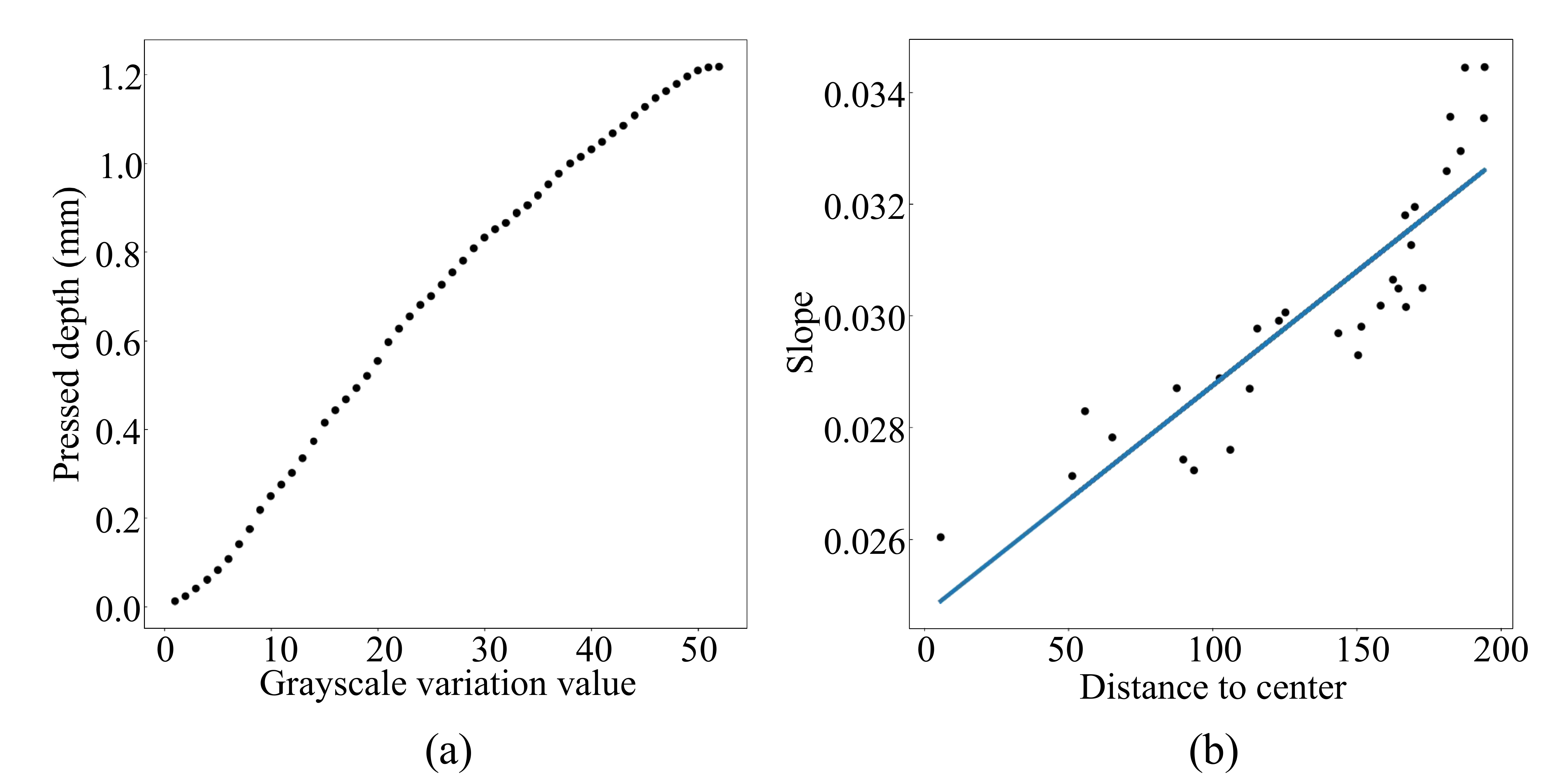}
	\caption{The calibration results. (a) The mapping list calibrated from a single image. (b) The fitted regression line for the linear regression method.}
	\label{fig:calibration}
\end{figure}

We calibrate the mapping list $\MappingList$ (shown in~\fig{fig:calibration} (a)) with only a single image because the ball has continuous geometry from center to edge. For the linear regression method, multiple images are required for fitting. \fig{fig:calibration} (b) shows the distribution of the collected points and the fitted regression line. The details about the collected images used to calibrate the parameters for two reconstruction methods are shown in~\tab{tab:images}. And we average images within a small time window to obtain stable and accurate calibration.

\begin{table}[ht]
\centering
\caption{Settings of the collected images for calibration and Reconstruction.}
\begin{tabular}{lcccc|}\toprule
 & \multicolumn{2}{c}{Calibration} & \multicolumn{1}{c}{\multirow{2}{*}{Reconstruction}} \\ \cmidrule(lr){2-3}
 & Single image & Linear regression & ~  \\ \midrule
Image amount    & 1       & 30            & 20                   \\
Ball radius    & 4.0 $mm$       & 4.0 $mm$            & 5.0 $mm$                       \\ 
Pressed place   & Near center       & Randomly            & Randomly                      \\ \bottomrule
\end{tabular}
\label{tab:images}
\end{table}

\subsection{Implementation Details of Reconstruction}
\label{sub:reconstruction_details}

In this section, we reconstruct the 3D geometry of the contact objects using the calibrated parameters. The captured images are first rectified with intrinsic and distortion coefficients to alleviate distortion. Next, we crop the image from $800\times600$ pixels to $580\times580$ pixels, aligning with the sensing field of $24\times24$$\texttt{ }mm^2$.

\begin{figure}[t]
	\centering
	\includegraphics[width= \linewidth]{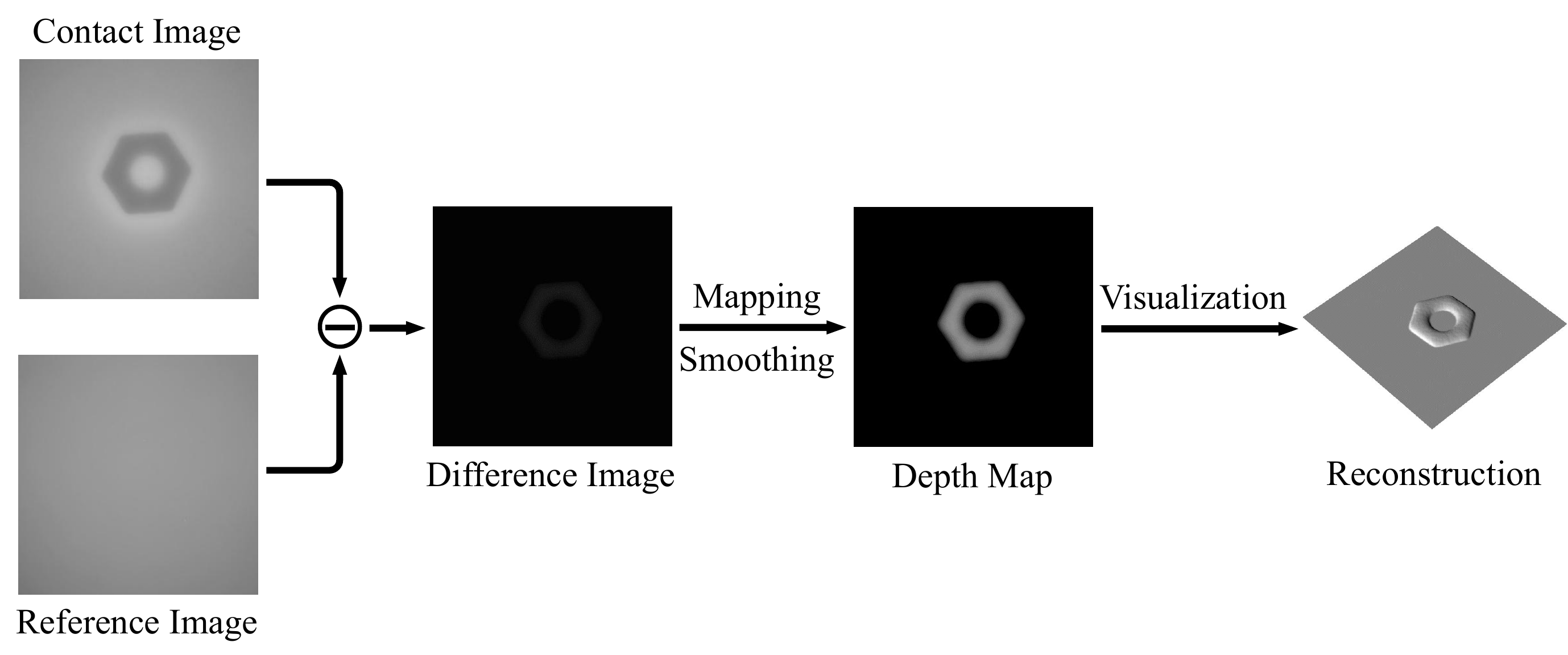}
	\caption{Pipeline of 3D shape reconstruction (a M4 nut is pressed as a showing example). Both the contact RGB image and reference RGB image have been rectified, cropped and converted to grayscaled images.}
	\label{fig:pipeline}
\end{figure}

\begin{figure}[t]
	\centering
	\includegraphics[width= \linewidth]{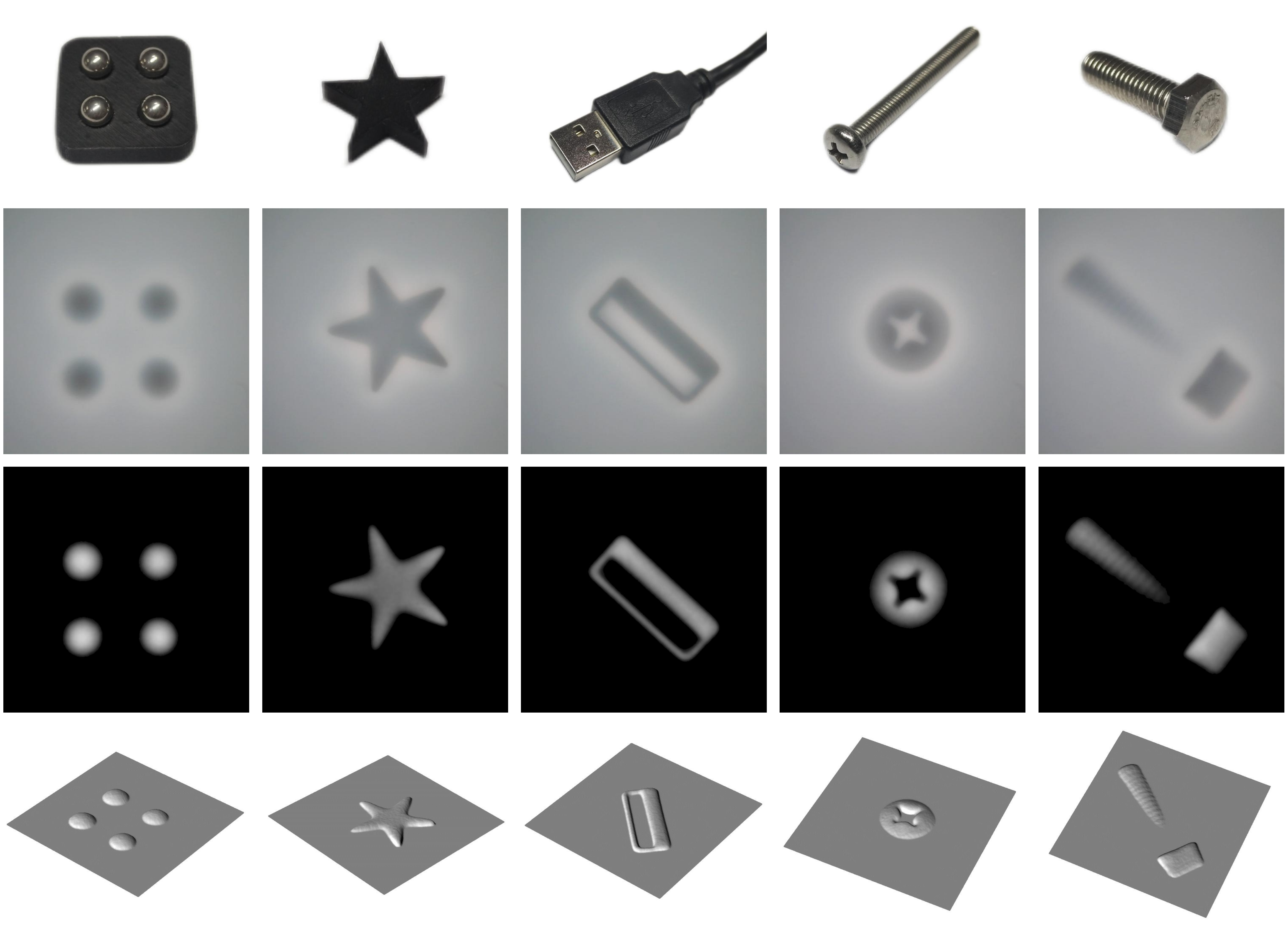}
	\caption{Reconstruction results with the single image method. From top row to bottom: visual images, sensor imprints, depth maps, and 3D reconstruction of a ball array, a 3D-printed star, a USB-mini cable head, a screw cap, and a M5 hexagonal screw.}
	\label{fig:showing}
\end{figure}

\fig{fig:pipeline} shows the pipeline to reconstruct 3D geometry of the contact objects. Firstly, we convert the contact image and reference image to the grayscale images. Then, we calculate the difference image by subtracting the reference image from the contact image. After that, the grayscale variation value $\ChangedValue$ is mapped to pressed depth $\PressedDepth$ for all pixels using the mapping list $\MappingList$ or regression parameters $k_c$ and $b_c$. We additionally apply two continuous Gaussian filters both with $7\times7$ kernel size to denoise the depth map. Finally, we convert the depth map to a point cloud to render and visualize the reconstructed objects.

The reconstruction algorithm is implemented with Python OpenCV~\cite{bradski2000opencv} and Open3D~\cite{zhou2018open3d} libraries. The algorithms for both two reconstruction methods are run on a desktop (Intel Core i7-8550U @1.80GHz) at about 20Hz without using a GPU. In practice, the rectifying, mapping, and smoothing processes take 8ms, 3ms, and 14ms respectively, while the visualization process takes up most of the time. In other words, we can get the final depth map from the raw captured image in 25ms, which is efficient for the downstream robotic tasks.

\subsection{Reconstruction Results}
\label{sub:reconstruction_results}

In order to quantitatively evaluate the reconstruction quality of the two methods, we press another metal ball of a different size to collect reconstruction test images. The details are summarized in the right-most column of~\tab{tab:images}. We compute the generated depth maps using the two reconstruction methods respectively and the actual depth maps by the same algorithms introduced in Section~\ref{sub:calibration}. Then we calculate the mean absolute error~(MAE) between the generated depth maps and corresponding actual depth maps. 

The test result in the standard setting is shown in \tab{tab:uniformity}. This result indicates that the single image method outperforms the linear regression method in the test images. There are two main reasons for the test results. On the one hand, \sensor{} diffuses light so that the reference image is with standard deviation~(std) as low as 4. This strongly improves the applicability to use the mapping list $\MappingList$ for all pixels. On the other hand, the errors from the fitting processes introduces additional errors for reconstruction. Therefore, the single image shape reconstruction method is chosen for \sensor{}.

\fig{fig:showing} shows the qualitative reconstruction results of some objects: a ball array, a 3D-printed star, a USB cable head, a screw cap and a M5 hexagonal screw. The reconstructed results are well aligned with the visual appearance of the objects.


\section{Experiments}
\label{sec:experiments}

In this section, we conduct experiments and aim to answer the following questions: 1) Is \sensor{} robust to various illumination conditions? 2) Can \sensor{} capture fine geometry of contact objects? 3) Can \sensor{} be extended to new sensors with non-planar contact surfaces?

\newcommand{\lengthled}{0.9 cm}
\newcommand{\lengthref}{0.9 cm}
\newcommand{\rownumber}{3}

\begin{table}[ht]
\centering
\caption{MAE ($mm$) on testing images using two reconstruction methods under different LED configurations}
\resizebox{\columnwidth}{!}{
\begin{tabular}{lccccc}\toprule
& Standard & Scheme 1 & Scheme 2 & Scheme 3 & Scheme 4  \\ 
& \includegraphics[width = \lengthled, height = \lengthled]{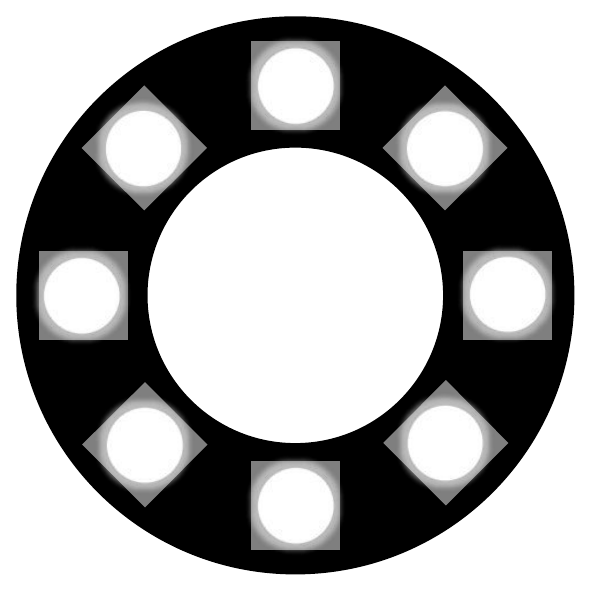}
& \includegraphics[width = \lengthled, height = \lengthled]{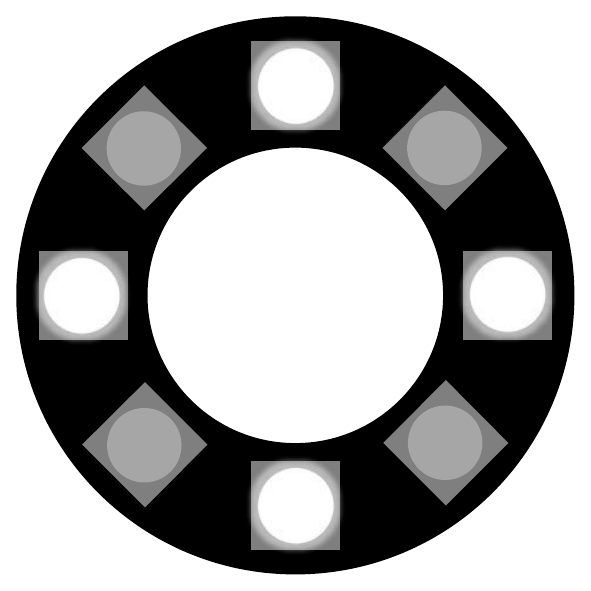} 
& \includegraphics[width = \lengthled, height = \lengthled]{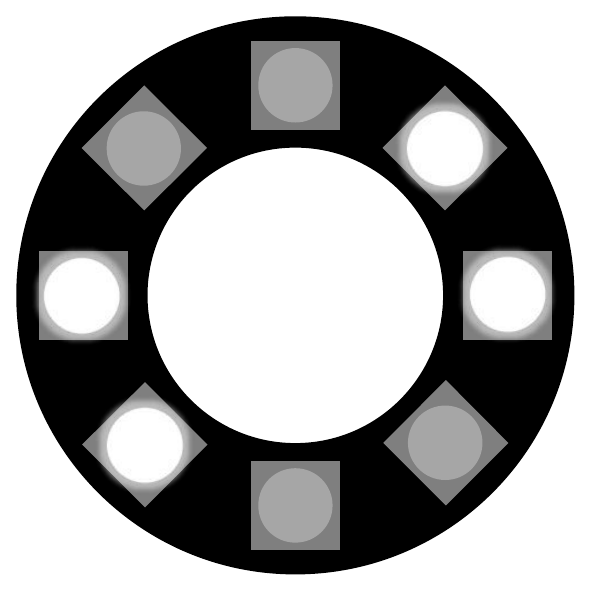}
& \includegraphics[width = \lengthled, height = \lengthled]{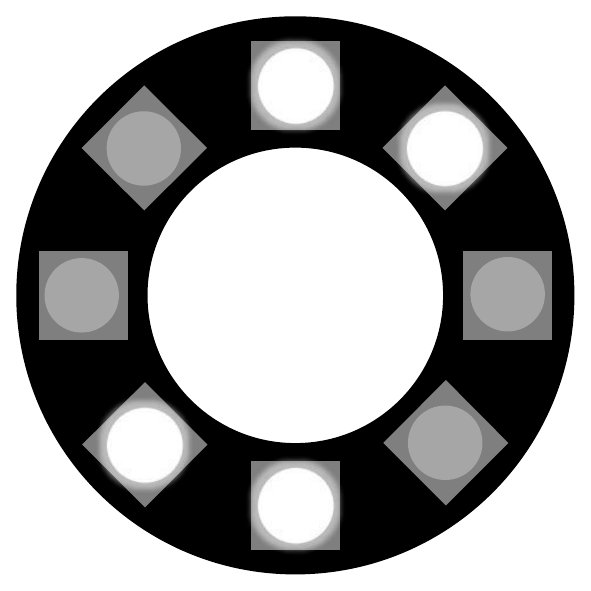} 
& \includegraphics[width = \lengthled, height = \lengthled]{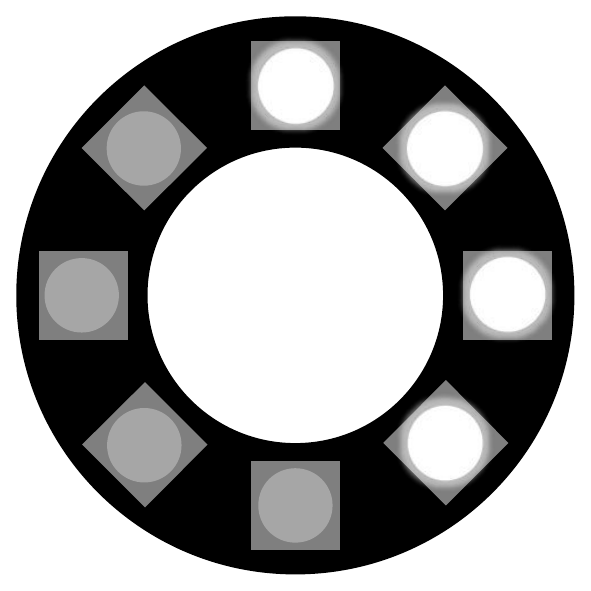}
\\ \midrule
Std of image      & 4.080     & 5.237     & 5.262     & 5.451      & 16.550        \\ 
\multicolumn{1}{l}{\multirow{\rownumber}{*}{Reference image}}
& \multirow{\rownumber}{*}{\includegraphics[width = \lengthref, height = \lengthref]{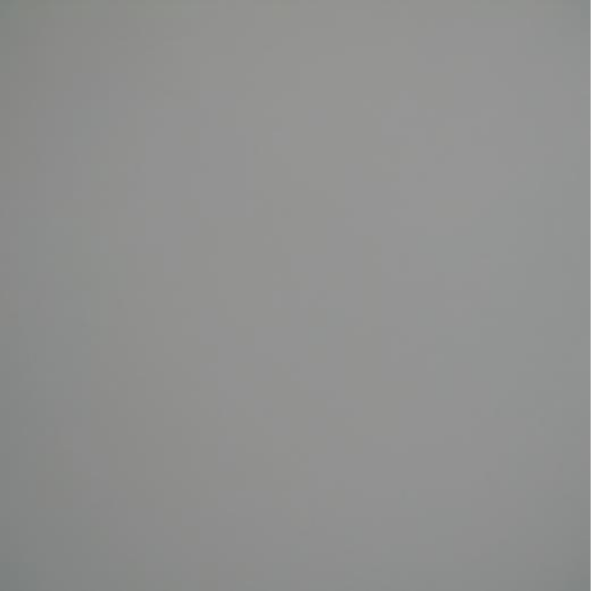}}
& \multirow{\rownumber}{*}{\includegraphics[width = \lengthref, height = \lengthref]{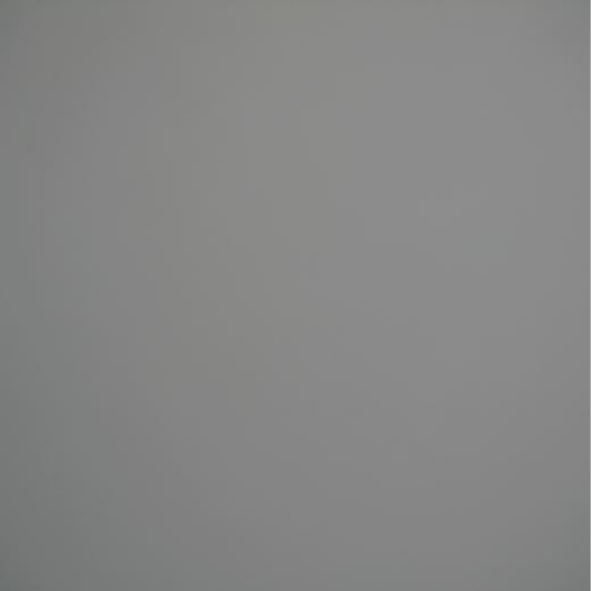} }
& \multirow{\rownumber}{*}{\includegraphics[width = \lengthref, height = \lengthref]{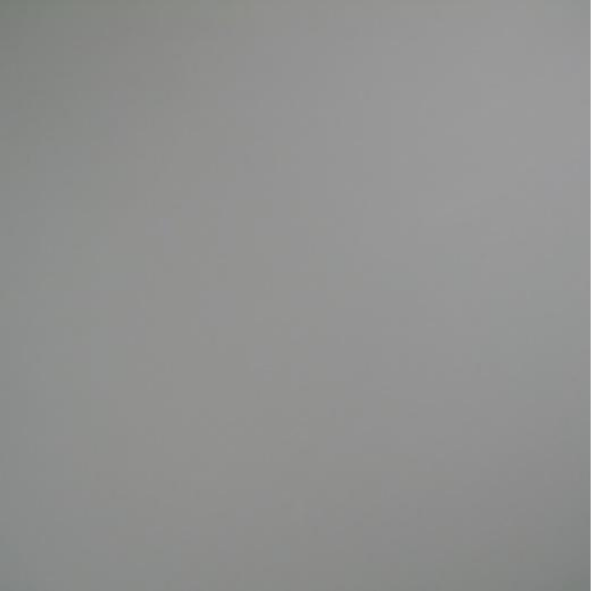}}
& \multirow{\rownumber}{*}{\includegraphics[width = \lengthref, height = \lengthref]{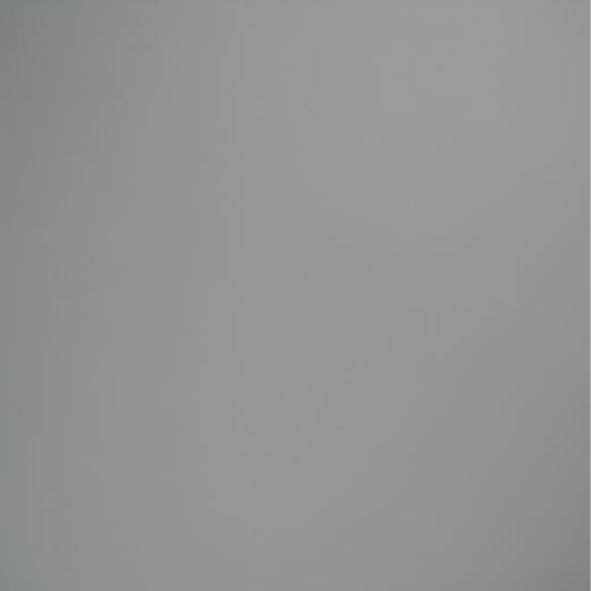} }
& \multirow{\rownumber}{*}{\includegraphics[width = \lengthref, height = \lengthref]{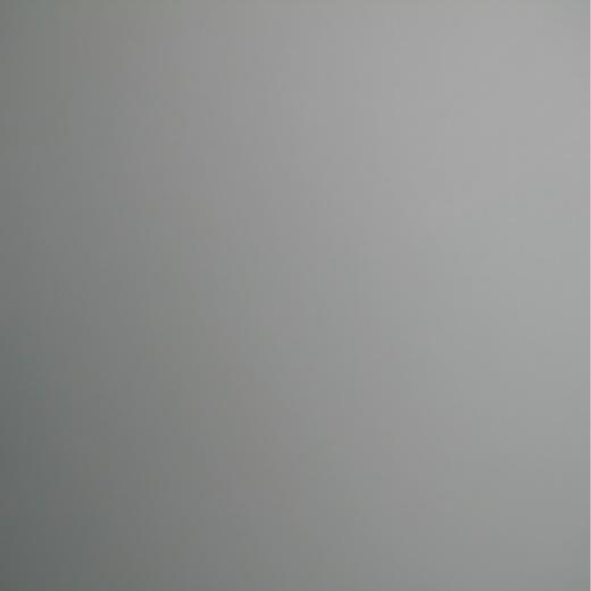}}
\\ 
\\
\\\midrule
Single image      & \textbf{0.0476}    & \textbf{0.0509}    & \textbf{0.0524}    & \textbf{0.0650}     & \textbf{0.0510}        \\ 
Linear regression  & 0.0534    & 0.0542    & 0.0586    & 0.0678     & 0.0651        \\
\bottomrule
\end{tabular}}
\label{tab:uniformity}
\end{table}

\subsection{Robustness to Illumination Conditions}
\label{sub:lighting}

Low dependency on illumination is the key advantage of \sensor{}. In this section, we change the illumination conditions in both uniformity and direction to evaluate whether \sensor{} is robust to illumination.

\myparagraph{Illumination uniformity.}
The LED ring used for \sensor{} has eight LEDs distributing evenly around a circle. As shown in~\tab{tab:uniformity}, we select four of them to form four new configurations, and repeat the processes of calibrations and reconstructions on test images for each of the new configurations. Specifically, for the linear regression method under the fourth configuration, we change the reference position $(u_c, v_c)$ in~\eq{center_fit} from center to the upper right corner, because the reference image tends to darken from this corner to the bottom left.

As \tab{tab:uniformity} shows, reconstruction with the single image method has the lowest MAE on test images under all LED configurations including the illumination that is relatively uneven~(e.g., the fourth configuration). This demonstrates that compared with the linear regression method that brings errors in fitting processes, the single image method has higher applicability in various illumination conditions. Besides, the MAE values of the single image method only increase slightly under all new LED configurations, which shows \sensor{}'s robustness to illumination uniformity.

\begin{figure}[t]
	\centering
	\includegraphics[width= \linewidth]{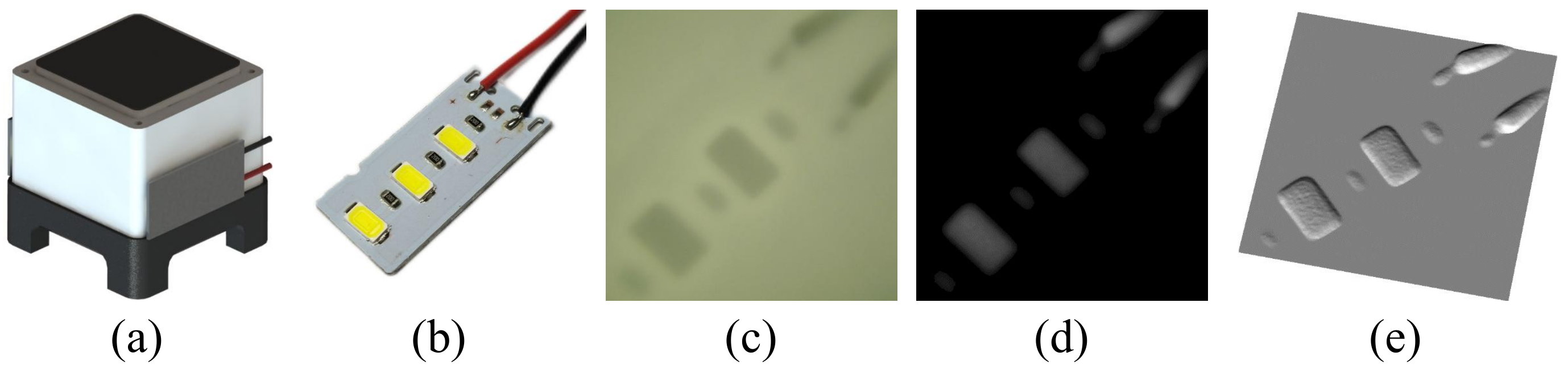}
	\caption{Lighting from the left and right sides. (a) The CAD model of \sensor{} lighting from the sides. (b) The LED used for illumination. (c) Raw image captured by \sensor{} when a same LED as (b) is pressed. (d) Depth map generated from the raw image. (e) 3D reconstruction of the LED.}
	\label{fig:illumination}
\end{figure}

\myparagraph{Illumination direction.}
There are two main illumination configurations for most vision-based tactile sensors: lighting from the bottom~\cite{sferrazza2019design} or from the sides~\cite{du2021high}. In this section, we test \sensor{} with LEDs on the sides as~\fig{fig:illumination} (a) shows. As shown in~\fig{fig:illumination} (b), We use the same type of LED as for illumination to press on \sensor{}'s surface. With the same pipeline, \fig{fig:illumination} (c), (d), and (e) show the tactile image, the depth map and the reconstructed shapes of the object respectively. These qualitative results show that \sensor{} maintains its reconstruction capability with the LEDs on the sides.

\subsection{Measuring Fine Geometry with Different Layer Thickness}
\label{sub:thickness}
\begin{figure}[t]
	\centering
	\includegraphics[width= \linewidth]{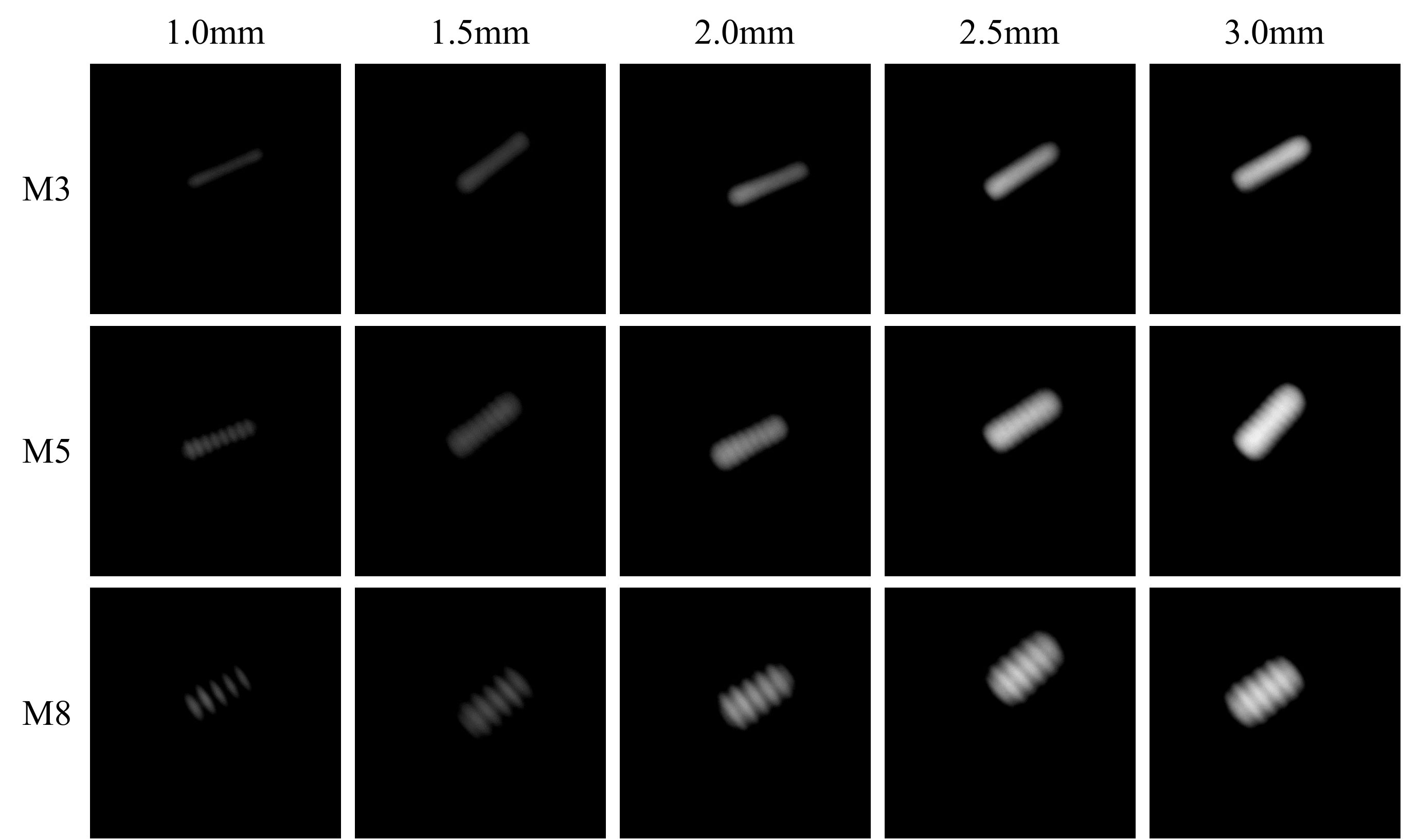}
	\caption{Depth maps generated from the sensors with different thicknesses of semitransparent layer when three sizes of set screws are pressed.}
	\label{fig:texture}
\end{figure}

To measure geometry with different fineness, we make semitransparent layers with different thicknesses, namely $1\texttt{mm}$, $1.5\texttt{mm}$, $2\texttt{mm}$, $2.5\texttt{mm}$, and $3\texttt{mm}$. Three different sizes of set screws are pressed to the sensors with these different layers. The depth maps shown in~\fig{fig:texture} indicate that sensors with thinner layers can capture finer geometry. However, the measure range of depth is bounded by the depth of a layer. Therefore, there is a trade-off between fineness and depth in measuring geometry.  
Applications on different scenarios could choose suitable layer thickness according to their measuring demands.

\subsection{Non-planar Surface}
\label{sub:surface}

\begin{figure}[t]
	\centering
	\includegraphics[width= \linewidth]{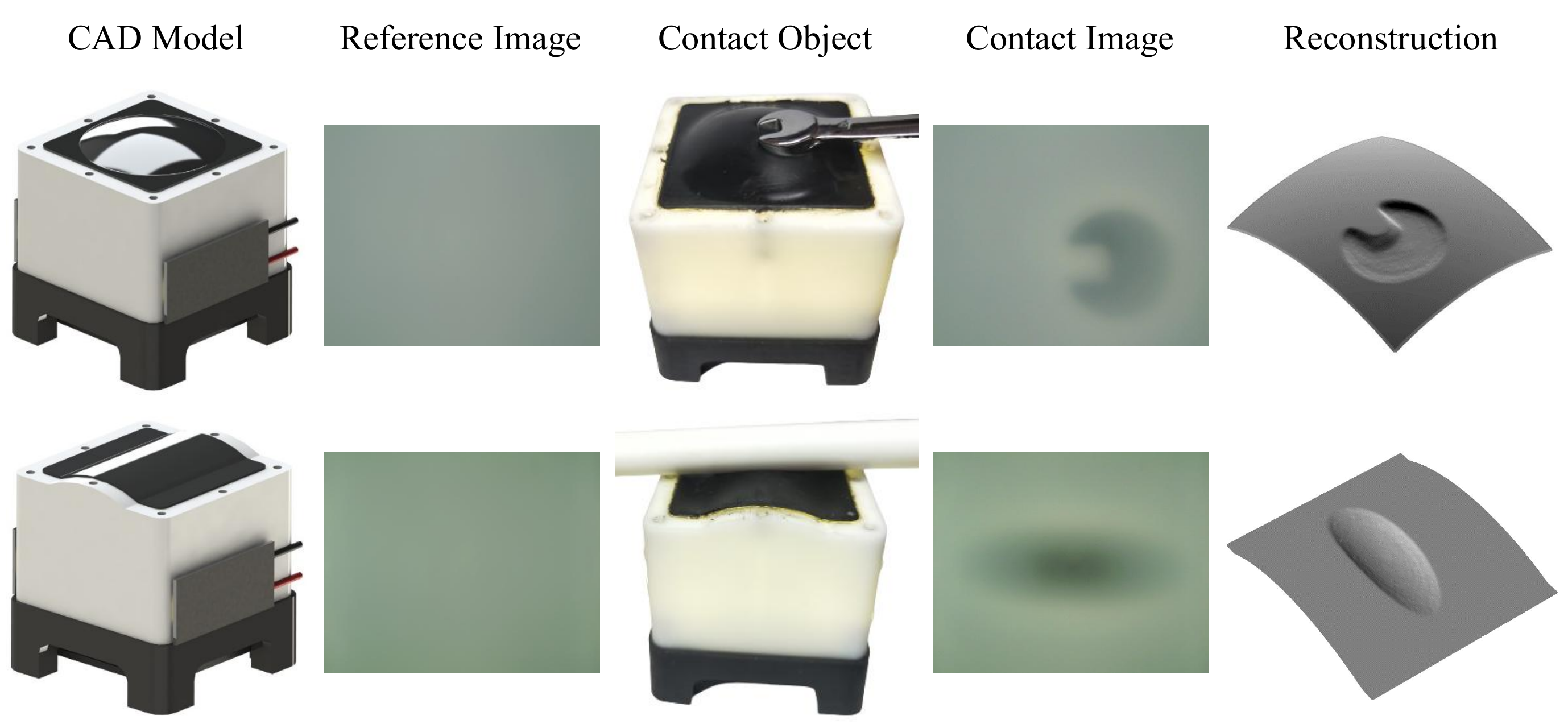}
	\caption{Two \sensor{} sensors equipped with non-planar surfaces, including sphere in the first row and cylinder in the second row.}
	\label{fig:non-planar}
\end{figure}

In this section, we extend \sensor{}'s surface to non-planar ones. We reshape the planar surface of \sensor{} into two non-planar surfaces: sphere and cylinder~(shown in~\fig{fig:non-planar}). Without changing the other parts, the reference images show that the illumination is still uniform with the help of the diffusion module. Therefore, the method to compute depth maps can be re-used for the sensors with non-planar surfaces. Both of the two sensors achieve to reconstruct their surfaces by projecting the depth maps to the non-planar surfaces with ray casting algorithm~\cite{do2022densetact}. The examples indicate that \sensor{} can perform reconstruction even when the surface is non-planar, which usually require complex mechanical design or intricate reconstruction algorithms for other sensors.

\section{Applications}
\label{sec:application}

In this section, we investigate how \sensor{} can be used for pose estimation and object recognition. This also shows the potential of \sensor{} for robotic applications.

\subsection{Pose Estimation}
\label{sub:pose}

\begin{figure}[t]
	\centering
	\includegraphics[width= \linewidth]{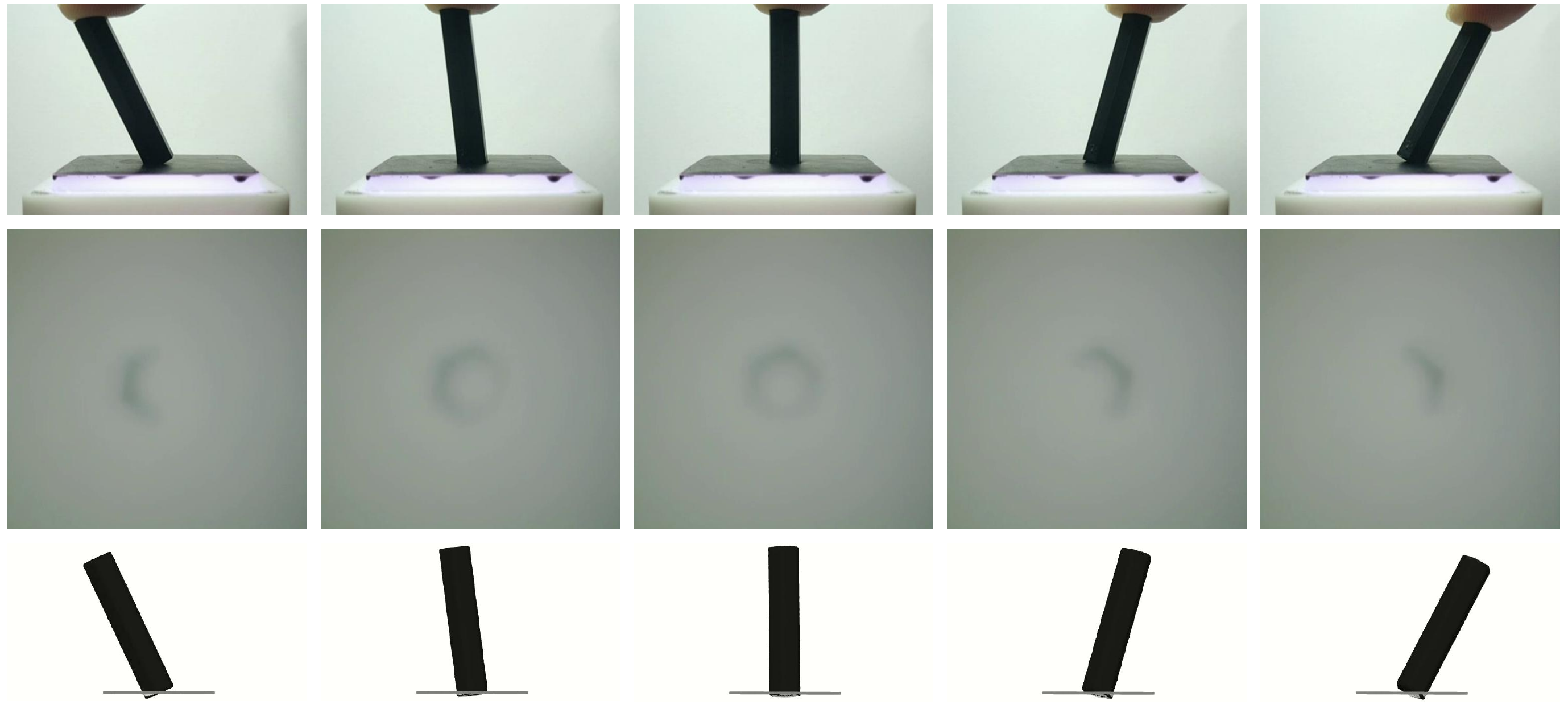}
	\caption{Pose tracking of a long M3 nut that rotates from the left around the back to the right. The first row shows the actual pose. The second row shows the images captured by \sensor{}. The third row shows the predicted pose. }
	\label{fig:pose}
\end{figure}

We apply Iterative Closest Point~(ICP)~\cite{besl1992method} on the reconstructed 3D point cloud for pose estimation. \fig{fig:pose} shows an example of tracking the pose of a long M3 nut that rotates from the left around the back to the right. The predicted poses are consistent with the actual poses. In practice, the ICP algorithm from Open3D~\cite{zhou2018open3d} runs at about 10 Hz on the same desktop introduced in Section~\ref{sub:reconstruction_details}.

\subsection{Object Recognition}

Object Recognition is important for downstream tasks such as distinguishing different types of machine components in a robot-based factory. We now introduce the datasets, the model, the training details, and the recognition performance. 

\myparagraph{Dataset.} We collect a total of 4,320 tactile images of objects within 12 different categories including metal balls, cylinders, nuts, bolts, spanners, legos of size 1x2 and 2x3, USB-mini cable heads, USB-C cable heads, keys, hex keys, and mesh pen holders. For all the objects, different pressures are applied to create images of different depths. 

\myparagraph{Architecture and training.} We use ResNet-18~\cite{he2016deep} as the architecture of our model. The dataset is randomly splitted to 70\% being training set and 30\% being test set, whereas a 4-fold cross validation is placed on the training set. We use cross-entropy loss and Adam optimizer with a learning rate scheduler, starting from 1e-3 and decaying by $0.3$ every 5 epoches. During training, we perform data augmentations including random crop, random flip, and random rotate. 

\myparagraph{Results.} After 15 epoches, the resnet-18 model achieves 96\% accuracy on the test set. This demonstrates that our \sensor{} sensor can capture important features for classifying different objects.


\section{Conclusion}
\label{sec:conclusion}

We propose \sensor{}, a robust, low-cost, and easy-to-fabricate tactile sensor that measures high-resolution 3D geometry accurately. The core contribution of \sensor{} is that it maintains tactile sensing performance while relieving a prevalent issue in previous sensors: heavy dependency on illumination.  Experiments show that \sensor{} not only measures geometry accurately under various illumination conditions, but also captures details of object with fine geometry. More importantly, \sensor{} can be easily extended with non-planar surfaces, achieving reconstruction efficiently. Finally, \sensor{} can be used for downstream tasks such as pose estimation and object recognition, exhibiting its potential to be used in real world scenarios.

\bibliographystyle{IEEEtran}
\bibliography{main}

\end{document}